# EAFP-Med: An Efficient Adaptive Feature Processing Module Based on Prompts for Medical Image Detection

Xiang Li[1, 2], Long Lan[1, 2], Husam Lahza[3], Shaowu Yang[1, 2], Shuihua Wang[4, 7], Wenjing Yang[1, 2, *], Hengzhu Liu[1, *], Yudong Zhang[3,5,6*]

1 College of Computer Science and Technology, National University of Defense Technology, Changsha 410073, P R China

2 Institute for Quantum Information & State Key Laboratory of High Performance Computing, College of Computer Science and Technology, National University of Defense Technology, Changsha 410073, P R China

3 Department of Information Technology, Faculty of Computing and Information Technology, King Abdulaziz University, Jeddah 21589, Saudi Arabia

4 Department of Biological Sciences, Xi'an Jiaotong-Liverpool University, Suzhou, Jiangsu 215123, China

5 School of Computing and Mathematical Sciences, University of Leicester, Leicester, LE1 7RH, UK

6 School of Computer Science and Engineering, Southeast University, Nanjing, Jiangsu 210096, China

7 Department of Mathematical Sciences, University of Liverpool, Liverpool, L69 3BX, UK.

E-mails: lixiang@nudt.edu.cn (Xiang Li), long.lan@nudt.edu.cn (Long Lan), hlahza@kau.edu.sa (Husa Lahza), shaowu.yang@nudt.edu.cn (Shaowu Yang) shuihuawang@ieee.org, shuihua.wang@xjtlu.edu.cn, (Shuihua Wang), wenjing.yang@nudt.edu.cn (Wenjing Yang), hengzhu_liu@263.net, hengzhuliu@nudt.edu.cn, (Hengzhu Liu), yudongzhang@ieee.org, yudongzhang@seu.edu.cn (Yudong Zhang)

* Correspondence should be addressed to Wenjing Yang, Hengzhu Liu, Yudong Zhang

**Abstract:** In the face of rapid advances in medical imaging, cross-domain adaptive medical image detection is challenging due to the differences in lesion representations across various medical imaging technologies. To address this issue, we draw inspiration from large language models to propose EAFP-Med, an efficient adaptive feature processing module based on prompts for medical image detection. EAFP-Med can efficiently extract lesion features of different scales from a diverse range of medical images based on prompts while being flexible and not limited by specific imaging techniques. Furthermore, it serves as a feature preprocessing module that can be connected to any model front-end to enhance the lesion features in input images. Moreover, we propose a novel adaptive disease detection model named EAFP-Med ST, which utilizes the Swin Transformer V2 – Tiny (SwinV2-T) as its backbone and connects it to EAFP-Med. We have compared our method to nine state-of-the-art methods. Experimental results demonstrate that EAFP-Med ST achieves the best performance on all three datasets (chest X-ray images, cranial magnetic resonance imaging images, and skin images). EAFP-Med can efficiently extract lesion features from various medical images based on prompts, enhancing the model's performance. This holds significant potential for improving medical image analysis and diagnosis.

## 1 Introduction

COVID-19 is a respiratory illness caused by severe acute respiratory syndrome coronavirus 2 (SARS-CoV-2). The most common symptoms are fever, cough, and fatigue. Most people who contract COVID-19 will recover without hospitalization. However, some individuals may develop severe illness that requires hospitalization and intensive care. Severe cases can lead to complications such as pneumonia, acute respiratory distress syndrome, and multi-organ failure. Since December 2019, more than 760 million cases of COVID-19 have been reported globally, resulting in over 6.9 million deaths, but the real numbers are thought to be much higher. As of June 2023, over 13 billion vaccine doses have been administered (WHO, 2023).

Monkeypox, a disease with symptoms similar to smallpox, is a zoonosis, a disease that is transmitted from animals to humans. It can be transmitted through contact with bodily fluids, skin lesions or internal mucosal surfaces (such as the mouth or throat), respiratory droplets, and contaminated objects (WHO, 2023). Since 1 January 2022, 115 Member States from all 6 World Health Organization (WHO) regions have reported cases of smallpox to WHO. As of 30 September 2023, a total of 91,123 laboratory-confirmed cases and 663 probable cases, including 157 deaths, have been reported to WHO (WHO, 2023).

Dementia is a syndrome that can be caused by a number of progressive diseases that affect memory, thinking, behavior, and the ability to perform daily activities. Alzheimer's disease is the most common type of dementia, accounting for 60-70% of cases. Alzheimer's disease and other forms of dementia are now among the top 10 causes of death worldwide. Women are disproportionally affected: Globally, 65% of deaths from Alzheimer's and other forms of dementia are women. Lack of diagnosis is a major problem. Even in high-income countries, only one-fifth to one-half of dementia cases are routinely recognized. Nearly 65.7 million people worldwide are expected to have dementia by 2030, and the number of people living with dementia is expected to increase from 50 million to 152 million by 2050 (WHO, 2012; WHO, 2017; WHO, 2020).

It can be seen that the current global health problems are still facing more challenges, and the diagnosis of diseases through the detection of medical images can effectively promote the development of medical work and alleviate the pressure on global health. Medical image detection is an important task involving identifying and classifying the lesion in the image. However, the distribution of lesions in medical images is also different due to the various medical imaging technologies used in the diagnosis of different diseases. Therefore, it is challenging to use a common method for efficient adaptive detection of medical images for multiple diseases.

To address this challenge, we propose an efficient feature adaptive processing module based on prompts for medical image detection. The module consists of multiple sub-modules (three feature extractors, three feature adaptors, and a prompt-based parameter adaptor). Feature extractors are responsible for extracting lesion features of varying scales, feature adaptors are used to enhance and fuse the extracted features, and the prompt-based parameter adaptor receives prompts and dynamically updates parameters that align with the target task for the module, ultimately providing more precise and robust lesion features to the backbone or downstream network. The highlights of this paper are as follows:

(i) An efficient adaptive feature processing module based on prompt is proposed for medical images detection, which is named EAFP-Med.

(ii) EAFP-Med can dynamically update parameters according to prompts, and adaptively extract lesion features from medical images generated by various medical imaging technologies.

(iii) EAFP-Med can be equipped at the front end of any backbone as a feature preprocessing module to provide more precise and robust lesion features to the downstream network.

(iv) We compared our method with nine state-of-the-art methods on three datasets, and experimental results show that our method can achieve the best performance on all three datasets.

The following sections of the paper are arranged as follows: Section 2 presents the related work of this study. Section 3 introduces the structure of EAFP-Med module. Section 4 presents the three datasets used in our experiments and experimental results. The paper is concluded in Section 5.

## 2 Related work

Many researchers are dedicated to advancing the development and application of AI-assisted diagnostic systems. In addition to leveraging large language models for medical knowledge question-answering, many are also employing various deep learning techniques to analyze medical images for disease diagnosis purposes.

Alsattar, et al. (2024) propose a novel dynamic location-based decision (DLBD) system that integrates four deep transfer learning models and twelve machine learning models. However, the complexity of this method lies in its extensive use of deep learning and machine learning models, which are accompanied by a significant number of parameters and high training costs. Seethi, et al. (2024) use mass spectrometry data for the diagnosis of COVID-19 and employed explainable artificial intelligence to explain the decision-making process at both the local (each sample) and global (all samples) levels. Experimental results show that the Cal RF+AD+PT method achieves the best accuracy of 94.12%. However, the authors did not utilize state-of-the-art deep learning methods, and the 94.12% detection accuracy alone does not fully demonstrate the method's effectiveness. Additionally, the authors only visualized the experimental data and did not use other more intuitive visualization techniques. Huo, et al. (2024) propose a three-branch hierarchical multi-scale feature fusion network structure called HiFuse, which utilizes the characteristics of convolution and self-attention-based Transformers to effectively fuse multi-scale global and local features. Experimental results show that HiFuse achieves good performance on three medical image datasets. However, this method focuses on extracting local and global features but overlooks the extraction of regional features in medical images. Therefore, the model's applicability still needs to be verified.

Bi, et al. (2023) propose a characteristic information aggregation hypergraph convolutional network (CIA-HGCN) based on hypergraph structure. This method achieves an accuracy of 88.3% in the task of Alzheimer's disease identification. Sharma, et al. (2023) use eight vanilla convolution blocks for feature extraction from fused input images, with a convolution kernel size of 7 in each block. Although the method achieves an accuracy of 97.33% in experiments, the relatively large receptive field of stacked convolution kernels makes it difficult to efficiently extract fine-grained local features. Lakhan, et al. (2023) propose an evolutionary deep convolutional neural network scheme (EDCNNS). Compared to existing research, EDCNNS reduced failure rates by 29% in different Alzheimer's disease categories and improved selection accuracy by 50%. Hu, et al. (2023) propose a model called Conv-Swinformer, which focuses on extracting local fine-grained features. However, while emphasizing the extraction of local features is crucial for Alzheimer's disease classification tasks, in other medical images, focusing solely on local features may not provide sufficient feature information. Therefore, the generalization and adaptive detection capabilities of Conv-Swinformer still need to be validated.

Akkilic, et al. (2024) propose a neural network with two hidden layers to analyze the numerical performance of a mathematical model for monkeypox transmission. However, due to the simplicity of the model, with only eleven and twenty-two neurons in the respective hidden layers, the performance of the model on the multi-classification tasks is limited, and the adaptive capability of the model is also constrained. Asif, et al. (2023) propose a metaheuristics optimization-based weighted average ensemble model (MO-WAE) for detecting monkeypox disease. This method achieves 97.78% accuracy. However, it requires high time cost to train the model optimized by metaheuristic algorithms, and the trained model can only show good performance for a specific task without good adaptive ability. Meanwhile, the three transfer learning models (DenseNet-201, MobileNet, and DenseNet-169) used in the experiments all contain a large number of parameters, which is also a noteworthy issue.

Thieme, et al. (2023) propose a deep convolutional neural network named MPXV-CNN to identify the skin lesions caused by MPXV. Currently, the method can be accessed through the Internet to provide some reference and guidance for the diagnosis of monkeypox disease. However, the sensitivity and specificity of the method for performing binary classification tasks (non-MPXV images, MPXV images) in the experiment are 0.83 and 0.91, and 0.965 and 0.898, respectively. We think that these results do not necessarily prove that this is a good monkeypox detection method, and the performance of MPXV-CNN in multi-classification tasks still needs to be validated.

To alleviate the above problems and achieve adaptive and efficient extraction of lesion features, we are inspired by the feature analysis of natural images (Li, et al., 2023) to divide the lesion region distribution in medical images into three types: global features, regional features, and local features. Global features refer to larger lesion areas in medical images, which usually span tens or hundreds of pixel values, mainly reflecting symmetry and multi-scale pattern repetition, texture similarity of lesions at the same scale, and lesion feature similarity and consistency. Regional features refer to moderate lesion areas in medical images, which usually span tens of pixel values, mainly reflecting a certain area of the lesion feature. Local features refer to smaller lesion areas in medical images, which usually span several or tens of pixel values, mainly reflecting the edge and color features of the lesion. The performance of the three feature regions in medical images is shown in Figure 1, where global features correspond to the blue rectangle areas in (a), regional features correspond to the red rectangle areas in (a), (b), and(c), local features correspond to the green rectangles areas in (b) and (c).

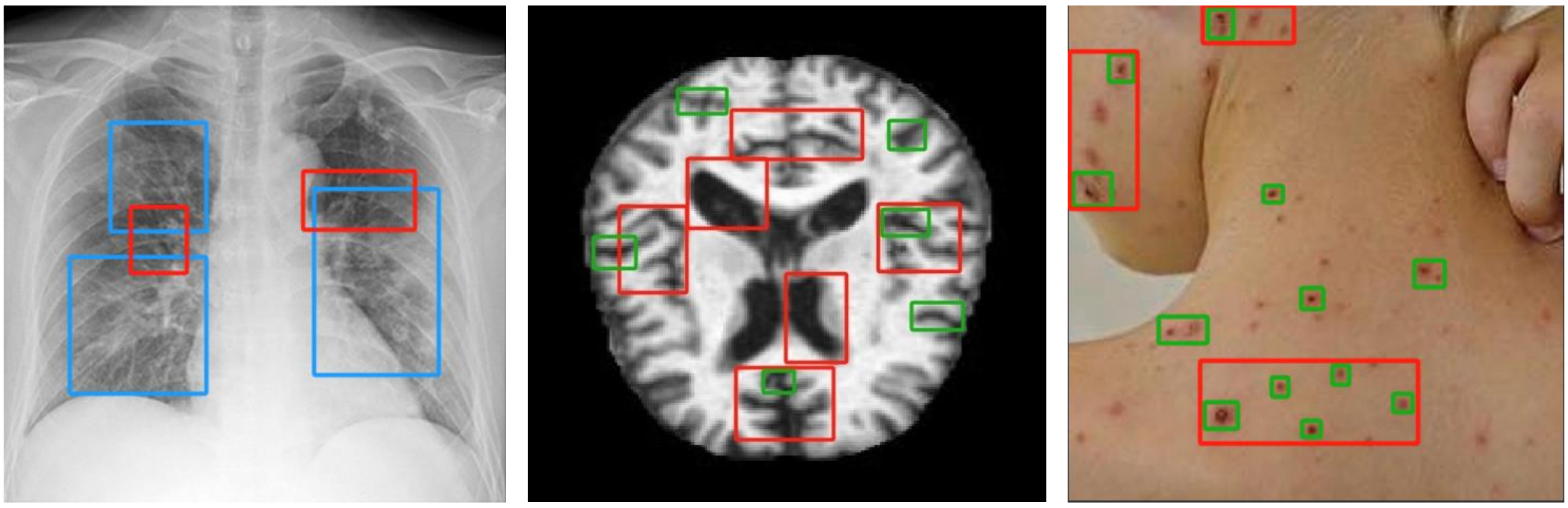

(a) A sample image of chest X-ray. (b) A sample image of cranial magnetic resonance imaging. (c) A sample image of Skin.

**Figure 1 Medical images show global (blue rectangles), regional (red rectangles), and local (green rectangles) features.**

## 3 EAFP-Med Module

In this section, we will introduce the structure of the EAFP-Med module. EAFP-Med consists of multiple sub-modules (three feature extractors, three feature adaptors, and a prompt-based parameter adaptor). The feature extractors are responsible for extracting the three types of lesion features. The feature adaptors work to effectively combine the extracted features. The prompt-based parameter adaptor is utilized to dynamically update the parameters of the EAFP-Med module based on prompts, ensuring optimal performance. A detailed illustration of the EAFP-Med module's structure can be found in Figure 2.

### 3.1 Feature Extractors

To effectively capture the three types of lesion features, we designed three feature extractors in EAFP-Med. The structure of all feature extractors is identical, consisting of several convolutional layers, batch norm layers, and Leaky ReLu layers. However, due to the different distributions of the target features they each need to process, we applied different parameter settings in the convolutional layers to ensure efficient extraction of the features.

EAFP-Med first passes the input image to the global feature extractor and the regional feature extractor for feature extraction. Subsequently, the local feature extractor performs local feature extraction on the extracted features from global feature extractor and regional feature extractor. Finally, all extracted features of the three types are fused to complete the feature preprocessing.

### 3.2 Feature Adaptors

The feature adaptors consist of convolutional layers with a convolution kernel size of 1, which is used to fuse feature maps from different feature extractors to enhance the features of lesions in the input image. The feature adaptor can effectively fuse three types of features, allowing for better overlaid with the original input image and feeding them to the model. Therefore, feature adaptors can bridge semantic and scale gaps when fusing multiple types of lesion features, and support the EAFP-Med to be added to the front-end of any backbone or downstream network, enhancing the module's versatility.

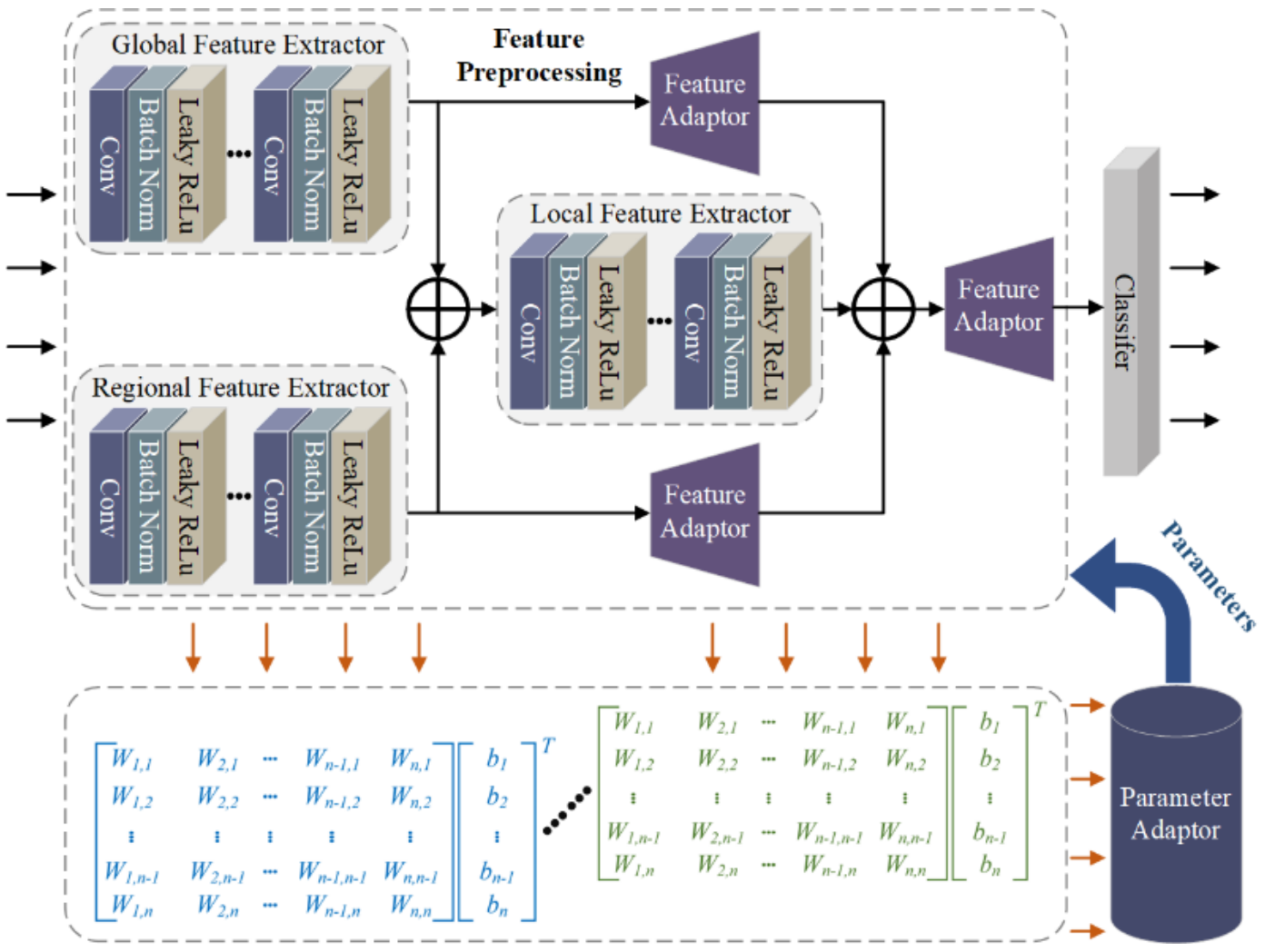

**Figure 2 The structure of the EAFP-Med module.**

## 3.3 Parameter Adaptor

The parameter adaptor is used to receive prompts and update parameters for EAFP-Med, and the parameter adapter acts like a pool storing sets of parameters that the model uses to perform various feature extraction tasks. The parameter sets in the parameter adaptor are obtained by pre-training the EAFP-Med on the target dataset. In our experiments, the parameter adaptor retains three sets of parameters obtained by pre-training the EAFP-Med on three target datasets. For specific experimental data, please refer to Section 4. Figure 3 illustrates our proposed adaptive medical image detection model named EAFP-Med ST, with EAFP-Med serving as the feature preprocessing module and SwinV2-T as the backbone.

# 4 Experimental Results

In this section, we first introduce the three datasets used in experiments. Next, we present the loss and accuracy curves of EAFP-Med and EAFP-Med ST during training and testing. Then, we display the confusion matrices and 17 measurement metrics for evaluating the performance of three methods (EAFP-Med, SwinV2-T, and EAFP-Med ST). Following this, we present the ROC curves to directly understand the classification performance of the models. Finally, we randomly selected one image from each of the datasets for visualizing Grad-Cam++, Cam-GB, and GB (Chattopadhay, et al., 2018) in order to intuitively display the attention regions of the models for different medical images.

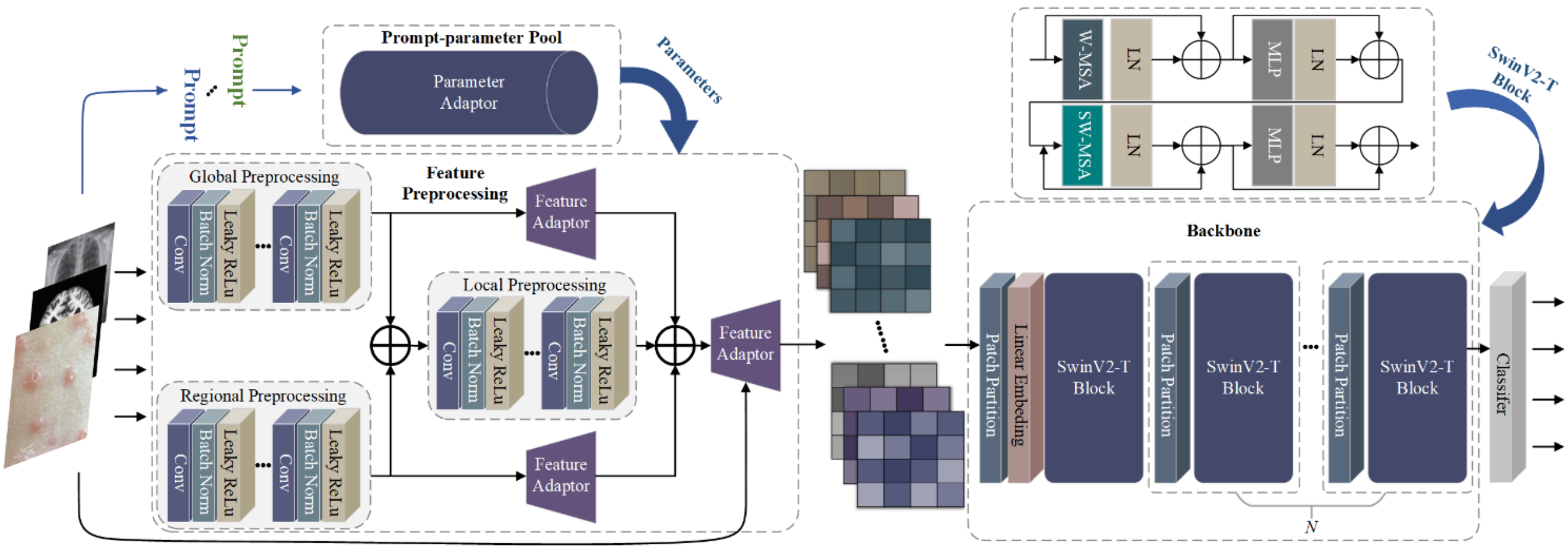

**Figure 3 The structure of our method.**

## 4.1 Datasets

To evaluate the adaptive detection performance of EAFP-Med, we conducted experiments on three datasets generated by different medical imaging technologies: chest X-ray image dataset (CXrI) (Asraf, 2021), cranial magnetic resonance imaging image dataset (CMRI) (Uraninjo, 2022), and skin image dataset (SK) (Bala, 2022). To ensure that the data distribution is as balanced as possible, we employ a range of data augmentation techniques such as random rotation, flipping, and cropping on these datasets. Here, CXrI consists of three image categories: COVID-19, normal, and pneumonia, with 2300 images in each category. The CMRI consists of four image categories: mild demented, moderate demented, non-demented, and very mild demented, with 4800 images in each category. CMRI also includes images from four categories: chickenpox, measles, monkeypox, and normal, with 1200 images in each category. Figure 4 to Figure 6 show sample images from each of the three datasets. For the above datasets, we randomly select 80% of the images as the training set and the remaining 20% as the testing set.

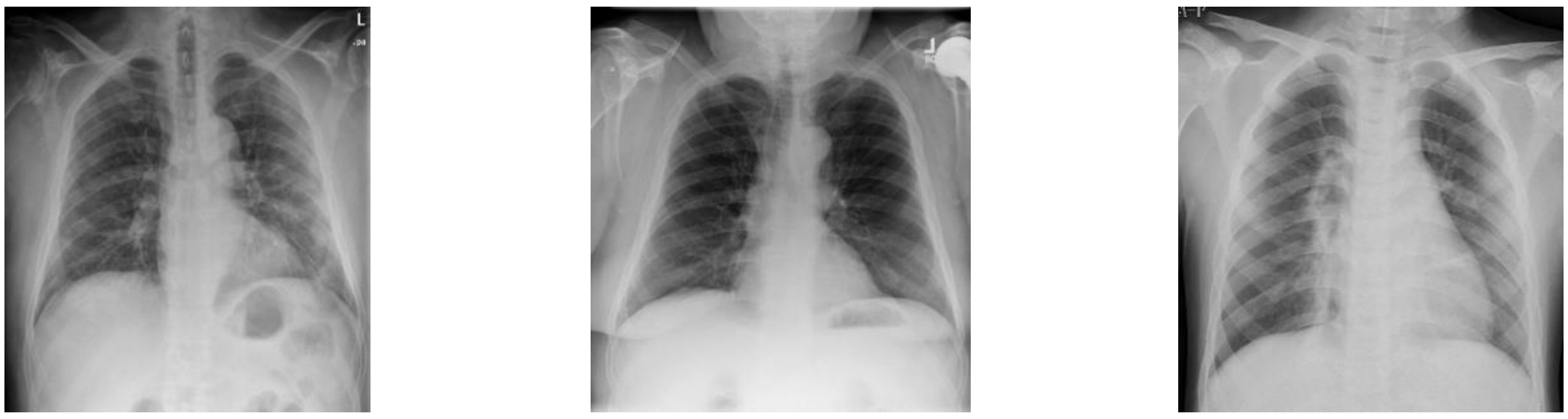

(a) A sample image of COVID-19. (b) A sample image of normal. (c) A sample image of pneumonia.

**Figure 4 Sample images from CXrI.**

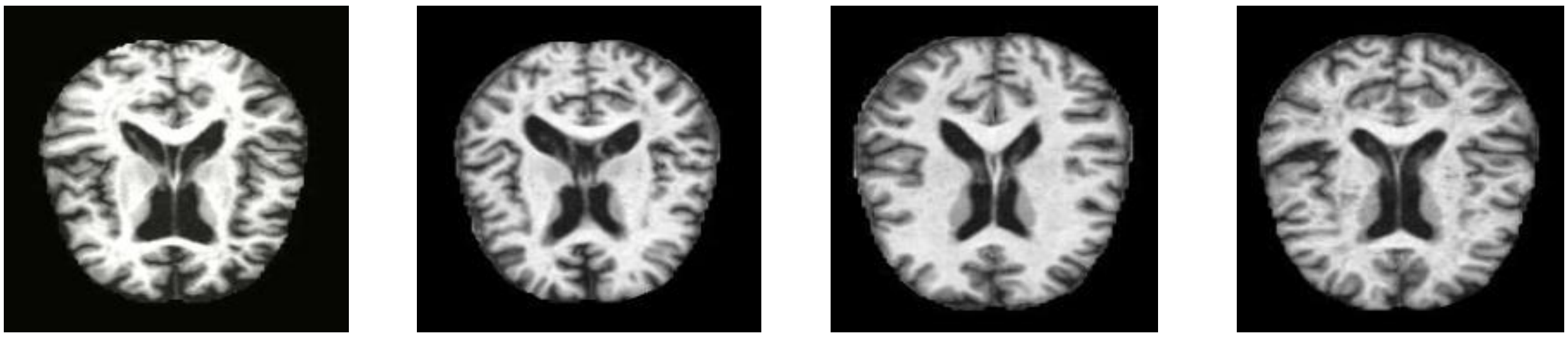

(a) A sample image of mild demented. (b) A sample image of Mild Demented. (c) A sample image of non-demented. (d) A sample image of very mild demented.

**Figure 5 Sample images from CMRI.**

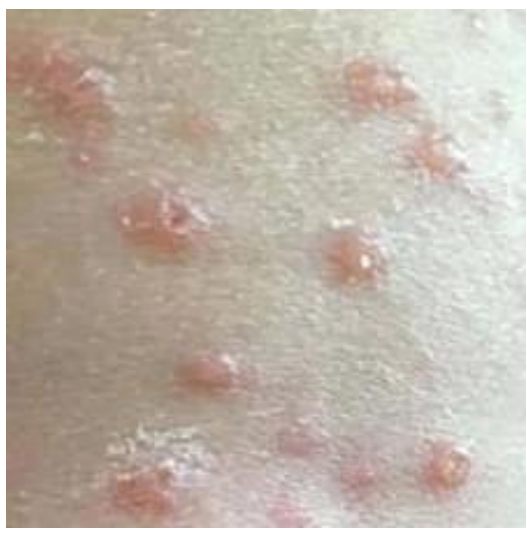

(a) A sample image of chickenpox.

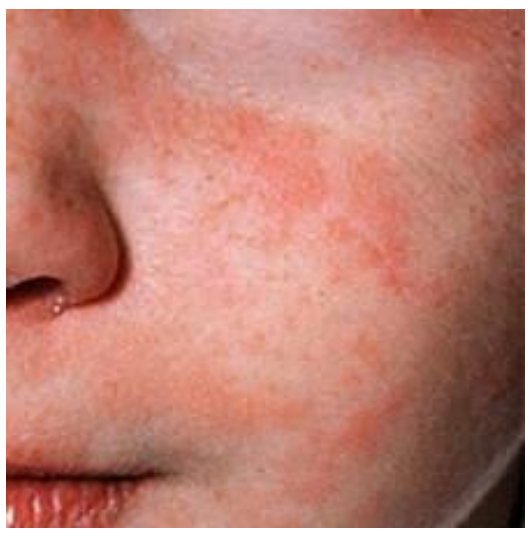

(b) A sample image of measles.

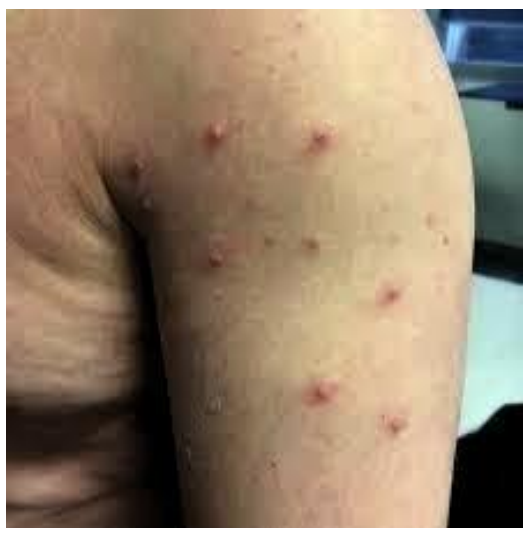

(c) A sample image of monkeypox.

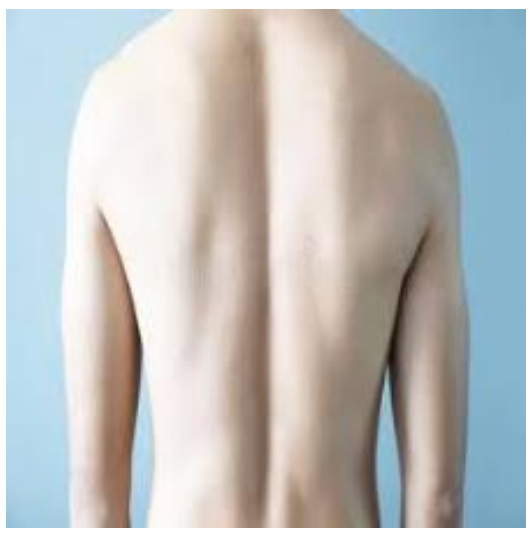

(d) A sample image of normal.

**Figure 6 Sample images from SK.**

## 4.2 Training our methods

Figure 7 shows the accuracy curves of the EAFP-Med module on three datasets, as well as the training and testing loss curves and accuracy curves of the EAFP-Med ST model on the same datasets.

EAFP-Med

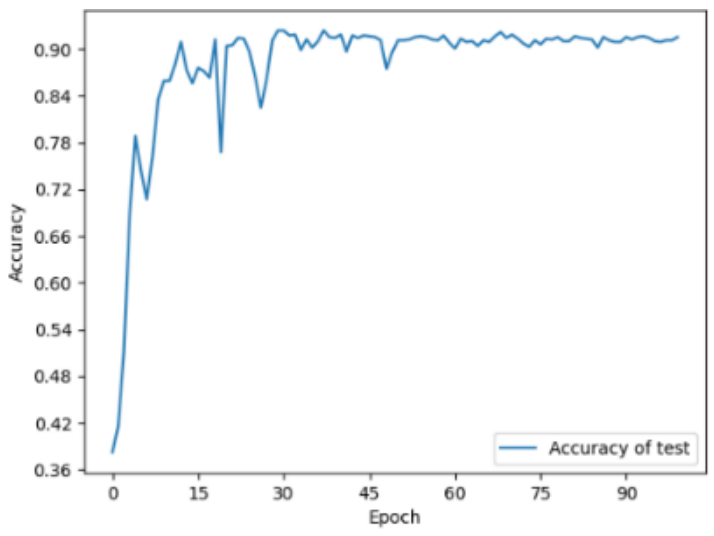

(a) Accuracy on CXrI.

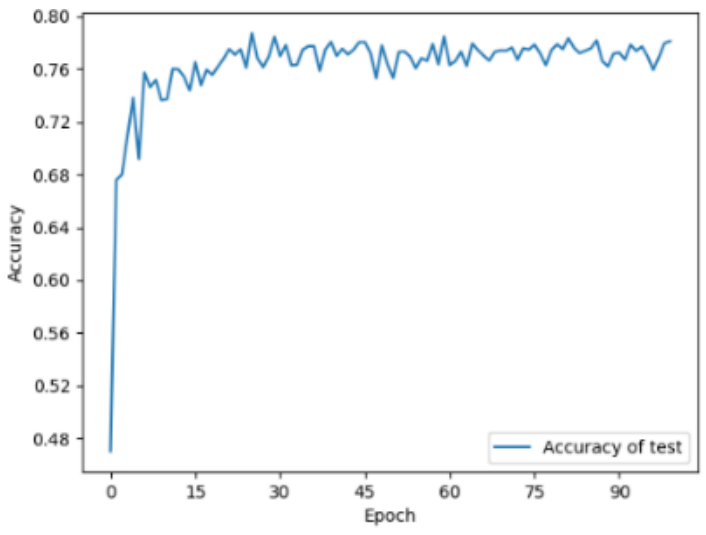

(b) Accuracy on CMRI.

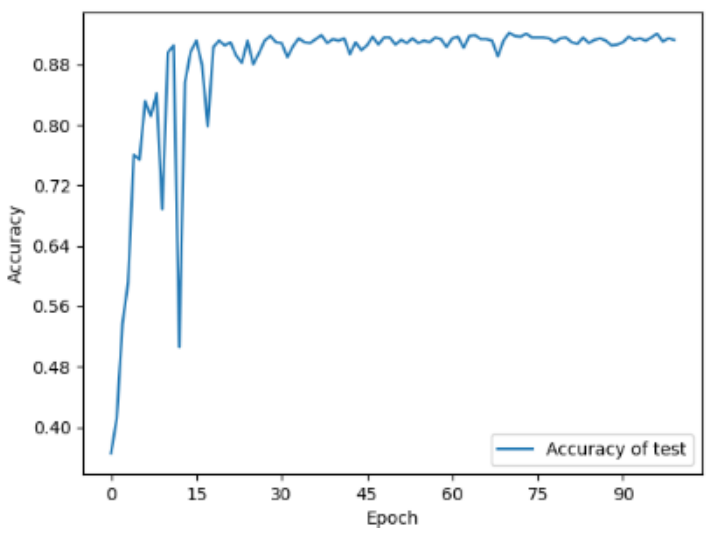

(c) Accuracy on SK.

EAFP-Med ST

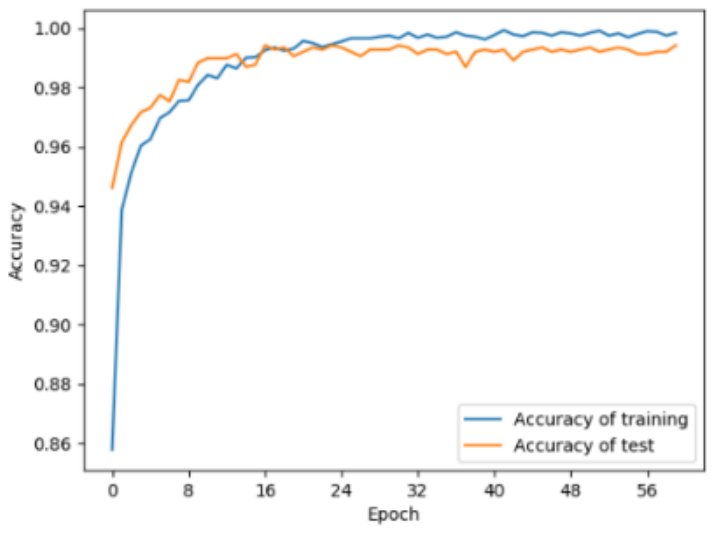

(d) Accuracies on CXrI.

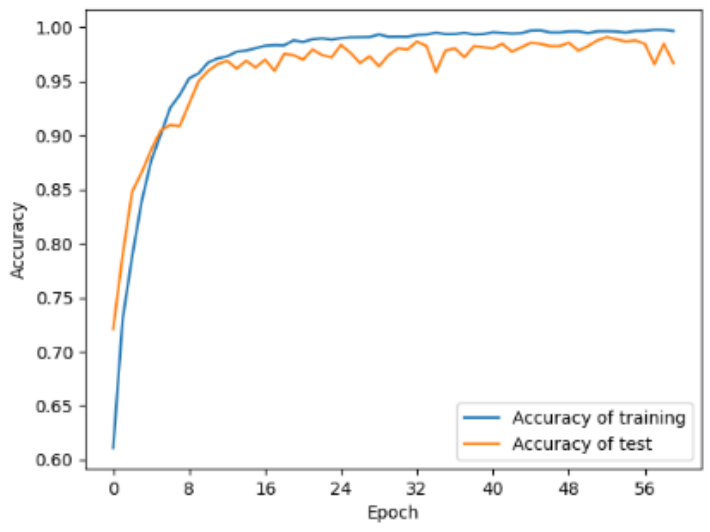

(e) Accuracies on CMRI.

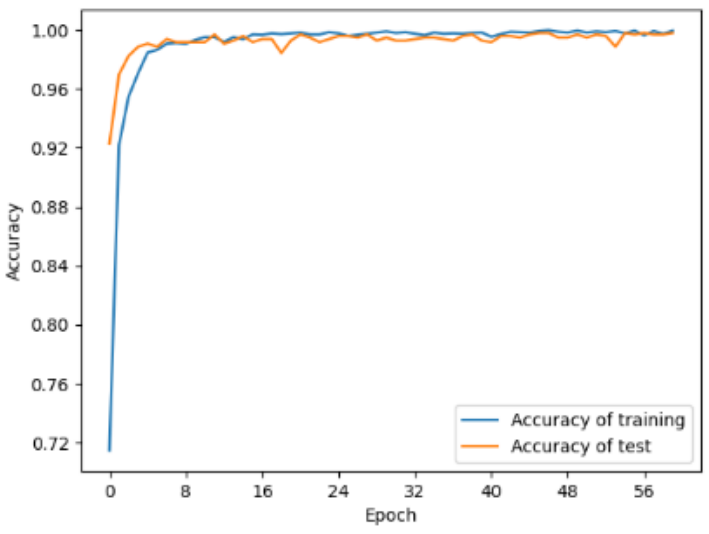

(f) Accuracies on SK.

EAFP-Med ST

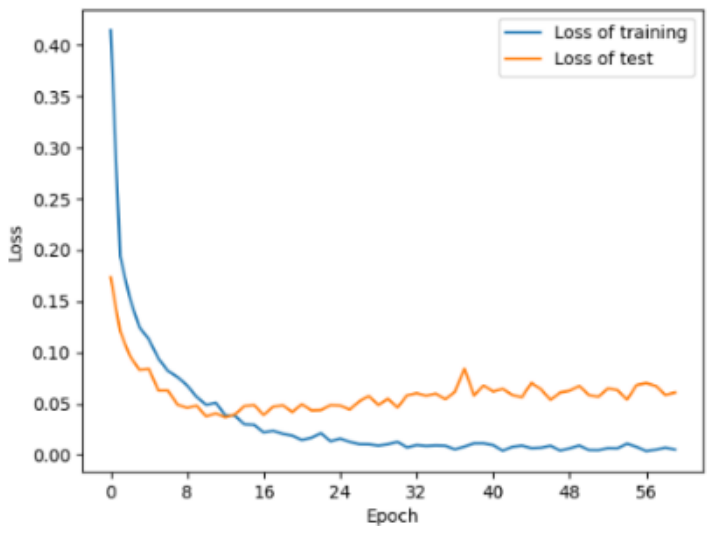

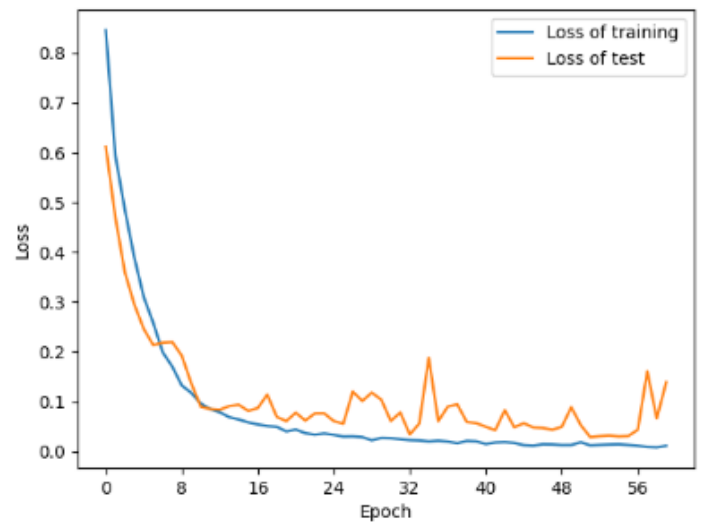

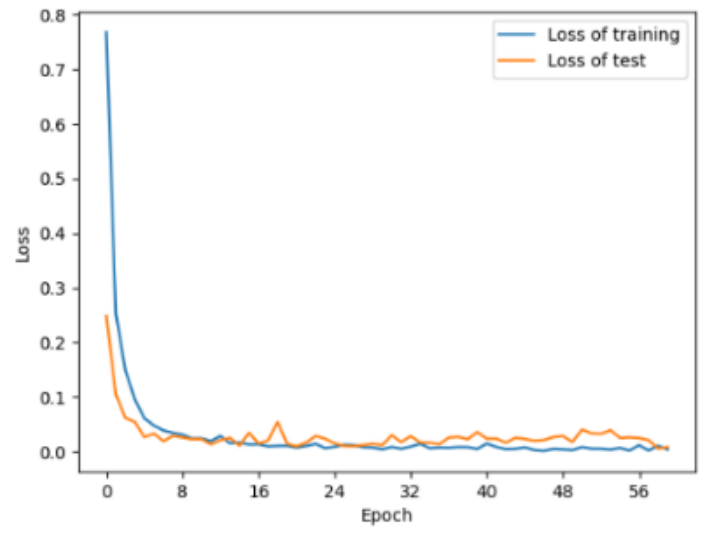

(g) Losses on CXrI. (h) Losses on CMRI. (i) Losses on SK.

**Figure 7 The accuracies and losses for the three methods.**

### 4.3 Confusion Matrix and Statistic Results

Table 1, Table 3, and Table 5, respectively, present the confusion matrices of three methods on three datasets. Here, COV denotes COVID-19 images in CXrI. Nor denotes normal images in CXrI. Pne denotes pneumonia images in CXrI. MD denotes mild demented images in CMRI. MoD denotes moderate demented images in CMRI. ND denotes non-demented images in CMRI. VMD denotes very mild demented images in CMRI. CP denotes chickenpox images in SK. Mea denotes measles images in SK. MP denotes monkeypox images in SK. Nor denotes normal images in SK.

Table 2, Table 4, and Table 6, respectively, present the statistical results of the confusion matrices. To comprehensively evaluate the performance of the models, we introduced the following 17 measurement metrics: accuracy (ACC), adjusted F-score (AGF), adjusted geometric mean (AGM), area under the ROC curve (AUC), Bray-Curtis dissimilarity (BCD), informedness or bookmaker informedness (BM), confusion entropy (CEN), F1 score (F1), false negative rate (FNR), false positive rate (FPR), G-measure geometric mean of precision and sensitivity (G), Matthews correlation coefficient (MCC), Otsuka-Ochiai coefficient (OOC), positive predictive value (PPV), random accuracy (RACC), receiver operating characteristic (ROC), true negative rate (TNR), true positive rate (TPR).

**Table 1 Confusion matrix for the detection of CXrI by the three methods.**

| Confusion Matrix | | Predicted Value | | | | | | | | |
|---|---|---|---|---|---|---|---|---|---|---|
| | | EAFP-Med | | | SwinV2-T | | | EAFP-Med ST | | |
| | | COV | Nor | Pne | COV | Nor | Pne | COV | Nor | Pne |
| Actual Value | COV | 395 | 54 | 11 | 448 | 12 | 0 | 460 | 0 | 0 |
| | Nor | 28 | 424 | 8 | 6 | 449 | 5 | 0 | 459 | 1 |
| | Pne | 11 | 65 | 384 | 2 | 51 | 407 | 0 | 12 | 448 |

**Table 2 Statistic results for the detection of CXrI by the three methods.**

| Statistic | EAFP-Med | | | SwinV2-T | | | EAFP-Med ST | | |
|---|---|---|---|---|---|---|---|---|---|
| Results | COV | Nor | Pne | COV | Nor | Pne | COV | Nor | Pne |
| ↑ ACC | 0.9246 | 0.8877 | 0.9312 | 0.9855 | 0.9464 | 0.958 | 1 | 0.9906 | 0.9906 |
| ↑ AGF | 0.9018 | 0.9137 | 0.8937 | 0.9817 | 0.965 | 0.9288 | 1 | 0.9948 | 0.984 |
| ↑ AGM | 0.9271 | 0.8858 | 0.9343 | 0.9861 | 0.9447 | 0.9607 | 1 | 0.9902 | 0.9914 |
| ↑ AUC | 0.9082 | 0.8962 | 0.9071 | 0.9826 | 0.9538 | 0.9397 | 1 | 0.9924 | 0.9864 |
| ↓ BCD | 0.0094 | 0.0301 | 0.0207 | 0.0015 | 0.0188 | 0.0174 | 0 | 0.004 | 0.004 |
| ↑ BM | 0.8163 | 0.7924 | 0.8141 | 0.9652 | 0.9076 | 0.8794 | 1 | 0.9848 | 0.9728 |
| ↓ CEN | 0.2786 | 0.3413 | 0.252 | 0.0744 | 0.1929 | 0.1512 | 0 | 0.0458 | 0.0466 |
| ↑ F1 | 0.8837 | 0.8455 | 0.8899 | 0.9782 | 0.9239 | 0.9335 | 1 | 0.986 | 0.9857 |
| ↓ FNR | 0.1413 | 0.0783 | 0.1652 | 0.0261 | 0.0239 | 0.1152 | 0 | 0.0022 | 0.0261 |
| ↓ FPR | 0.0424 | 0.1294 | 0.0207 | 0.0087 | 0.0685 | 0.0054 | 0 | 0.013 | 0.0011 |
| ↑ G | 0.884 | 0.8484 | 0.8919 | 0.9781 | 0.9251 | 0.9349 | 1 | 0.9861 | 0.9858 |
| ↑ MCC | 0.8288 | 0.7646 | 0.8441 | 0.9673 | 0.8857 | 0.9058 | 1 | 0.9791 | 0.9788 |
| ↑ OOC | 0.884 | 0.8484 | 0.8919 | 0.9781 | 0.9251 | 0.9349 | 1 | 0.9861 | 0.9858 |
| ↑ PPV | 0.9101 | 0.7809 | 0.9529 | 0.9825 | 0.877 | 0.9879 | 1 | 0.9745 | 0.9978 |
| ↓ RACC | 0.1048 | 0.1312 | 0.0973 | 0.1101 | 0.1237 | 0.0995 | 0 | 0.1138 | 0.1085 |
| ↑ TNR | 0.9576 | 0.8707 | 0.9794 | 0.9913 | 0.9315 | 0.9946 | 1 | 0.987 | 0.9989 |
| ↑ TPR | 0.8587 | 0.9217 | 0.8348 | 0.9739 | 0.9761 | 0.8848 | 1 | 0.9978 | 0.9739 |

**Table 3 Confusion matrix for the detection of CMRI by the three methods.**

| Confusion Matrix | | Predicted Value | | | | | | | | | | | |
|---|---|---|---|---|---|---|---|---|---|---|---|---|---|
| | | EAFP-Med | | | | SwinV2-T | | | | EAFP-Med ST | | | |
| | | MD | MoD | ND | VMD | MD | MoD | ND | VMD | MD | MoD | ND | VMD |
| Actual Value | MD | 745 | 72 | 230 | 153 | 1180 | 0 | 4 | 16 | 1174 | 0 | 8 | 18 |
| | Mod | 65 | 1006 | 83 | 46 | 0 | 1199 | 1 | 0 | 0 | 1200 | 0 | 0 |
| | ND | 96 | 25 | 945 | 134 | 29 | 0 | 1060 | 111 | 9 | 0 | 1148 | 43 |
| | VMD | 171 | 50 | 411 | 568 | 48 | 0 | 38 | 1114 | 7 | 0 | 30 | 1163 |

**Table 4 Statistic results for the detection of CMRI by the three methods.**

| Statistic Results | EAFP-Med | | | | SwinV2-T | | | | EAFP-Med ST | | | |
|---|---|---|---|---|---|---|---|---|---|---|---|---|
| | MD | MoD | ND | VMD | MD | MoD | ND | VMD | MD | MoD | ND | VMD |
| ↑ ACC | 0.836 | 0.929 | 0.796 | 0.799 | 0.98 | 0.9998 | 0.9619 | 0.9556 | 0.9913 | 1 | 0.9813 | 0.9796 |
| ↑ AGF | 0.7484 | 0.8956 | 0.8071 | 0.6511 | 0.9826 | 0.9996 | 0.9319 | 0.9475 | 0.9866 | 1 | 0.9725 | 0.9768 |
| ↑ AGM | 0.818 | 0.9235 | 0.7956 | 0.7634 | 0.98 | 0.9998 | 0.9573 | 0.9542 | 0.9906 | 1 | 0.98 | 0.9791 |
| ↑ AUC | 0.7643 | 0.8988 | 0.7932 | 0.6904 | 0.981 | 0.9996 | 0.9357 | 0.9465 | 0.9869 | 1 | 0.9731 | 0.9761 |
| ↓ BCD | 0.0128 | 0.0049 | 0.0489 | 0.0312 | 0.0059 | 0.0001 | 0.0101 | 0.0043 | 0.001 | 0 | 0.0015 | 0.0025 |
| ↑ BM | 0.5286 | 0.7975 | 0.5864 | 0.3808 | 0.9619 | 0.9992 | 0.8714 | 0.8931 | 0.9739 | 1 | 0.9461 | 0.9522 |
| ↓ CEN | 0.5311 | 0.2963 | 0.4919 | 0.5929 | 0.0963 | 0.0018 | 0.1582 | 0.1761 | 0.0525 | 0 | 0.0935 | 0.1 |
| ↑ F1 | 0.6544 | 0.8551 | 0.6588 | 0.5407 | 0.9605 | 0.9996 | 0.9205 | 0.9127 | 0.9824 | 1 | 0.9623 | 0.9596 |
| ↓ FNR | 0.3792 | 0.1617 | 0.2125 | 0.5267 | 0.0167 | 0.0008 | 0.1167 | 0.0717 | 0.0217 | 0 | 0.0433 | 0.0308 |
| ↓ FPR | 0.0922 | 0.0408 | 0.2011 | 0.0925 | 0.0214 | 0 | 0.0119 | 0.0353 | 0.0044 | 0 | 0.0106 | 0.0169 |
| ↑ G | 0.6553 | 0.8553 | 0.6678 | 0.5463 | 0.9608 | 0.9996 | 0.9214 | 0.9129 | 0.9824 | 1 | 0.9623 | 0.9596 |
| ↑ MCC | 0.5487 | 0.8083 | 0.5332 | 0.4223 | 0.9474 | 0.9994 | 0.8969 | 0.8832 | 0.9766 | 1 | 0.9498 | 0.946 |
| ↑ OOC | 0.6553 | 0.8553 | 0.6678 | 0.5463 | 0.9608 | 0.9996 | 0.9214 | 0.9129 | 0.9824 | 1 | 0.9623 | 0.9596 |
| ↑ PPV | 0.6917 | 0.8725 | 0.5662 | 0.6304 | 0.9387 | 1 | 0.961 | 0.8977 | 0.9866 | 1 | 0.968 | 0.9502 |
| ↓ RACC | 0.0561 | 0.0601 | 0.0869 | 0.0469 | 0.0655 | 0.0625 | 0.0575 | 0.0646 | 0.062 | 0.0625 | 0.0618 | 0.0638 |
| ↑ TNR | 0.9077 | 0.9592 | 0.7988 | 0.9075 | 0.9786 | 1 | 0.9881 | 0.9647 | 0.9956 | 1 | 0.9894 | 0.9831 |
| ↑ TPR | 0.6208 | 0.8383 | 0.7875 | 0.4733 | 0.9833 | 0.9992 | 0.8833 | 0.9283 | 0.9783 | 1 | 0.9567 | 0.9692 |

**Table 5 Confusion matrix for the detection of SK by the three methods.**

| Confusion Matrix | | Predicted Value | | | | | | | | | | | |
|---|---|---|---|---|---|---|---|---|---|---|---|---|---|
| | | EAFP-Med | | | | SwinV2-T | | | | EAFP-Med ST | | | |
| | | CP | Mea | MP | Nor | CP | Mea | MP | Nor | CP | Mea | MP | Nor |
| Actual Value | CP | 154 | 26 | 38 | 22 | 236 | 2 | 2 | 0 | 239 | 0 | 1 | 0 |
| | Mea | 28 | 185 | 12 | 15 | 3 | 233 | 1 | 3 | 1 | 238 | 0 | 1 |
| | MP | 16 | 19 | 190 | 15 | 10 | 4 | 225 | 1 | 1 | 3 | 235 | 1 |
| | Nor | 10 | 12 | 17 | 201 | 1 | 1 | 1 | 237 | 1 | 0 | 0 | 239 |

**Table 6 Statistic results for the detection of SK by the three methods.**

| Statistic Results | EAFP-Med | | | | SwinV2-T | | | | EAFP-Med ST | | | |
|---|---|---|---|---|---|---|---|---|---|---|---|---|
| | CP | Mea | MP | Nor | CP | Mea | MP | Nor | CP | Mea | MP | Nor |
| ↑ ACC | 0.8542 | 0.8833 | 0.8781 | 0.9052 | 0.9813 | 0.9854 | 0.9802 | 0.9927 | 0.9958 | 0.9948 | 0.9938 | 0.9969 |
| ↑ AGF | 0.7674 | 0.8427 | 0.8495 | 0.8832 | 0.9834 | 0.9805 | 0.9641 | 0.9911 | 0.9961 | 0.9939 | 0.9883 | 0.9967 |
| ↑ AGM | 0.8367 | 0.8761 | 0.8729 | 0.9013 | 0.9814 | 0.9847 | 0.9779 | 0.9925 | 0.9958 | 0.9946 | 0.993 | 0.9968 |
| ↑ AUC | 0.7833 | 0.8458 | 0.8493 | 0.8826 | 0.9819 | 0.9806 | 0.966 | 0.991 | 0.9958 | 0.9938 | 0.9889 | 0.9965 |
| ↓ BCD | 0.0167 | 0.001 | 0.0089 | 0.0068 | 0.0052 | 0 | 0.0057 | 0.0005 | 0.001 | 0.0005 | 0.0021 | 0.0005 |
| ↑ BM | 0.5667 | 0.6917 | 0.6986 | 0.7653 | 0.9639 | 0.9611 | 0.9319 | 0.9819 | 0.9917 | 0.9875 | 0.9778 | 0.9931 |
| ↓ CEN | 0.5021 | 0.4142 | 0.4147 | 0.3555 | 0.0939 | 0.0848 | 0.1034 | 0.0463 | 0.0286 | 0.032 | 0.0395 | 0.0215 |
| ↑ F1 | 0.6875 | 0.7676 | 0.7646 | 0.8154 | 0.9633 | 0.9708 | 0.9595 | 0.9855 | 0.9917 | 0.9896 | 0.9874 | 0.9938 |
| ↓ FNR | 0.3583 | 0.2291 | 0.2083 | 0.1625 | 0.0167 | 0.0292 | 0.0625 | 0.0125 | 0.0042 | 0.0083 | 0.0208 | 0.0042 |
| ↓ FPR | 0.075 | 0.0792 | 0.0931 | 0.0722 | 0.0194 | 0.0097 | 0.0056 | 0.0056 | 0.0042 | 0.0042 | 0.0014 | 0.0028 |
| ↑ G | 0.6893 | 0.7676 | 0.765 | 0.8157 | 0.9635 | 0.9708 | 0.9598 | 0.9855 | 0.9917 | 0.9896 | 0.9874 | 0.9938 |
| ↑ MCC | 0.5956 | 0.6898 | 0.6832 | 0.7522 | 0.951 | 0.9611 | 0.9469 | 0.9806 | 0.9889 | 0.9861 | 0.9833 | 0.9917 |
| ↑ OOC | 0.6893 | 0.7676 | 0.765 | 0.8157 | 0.9635 | 0.9708 | 0.9598 | 0.9855 | 0.9917 | 0.9896 | 0.9874 | 0.9938 |
| ↑ PPV | 0.7404 | 0.7645 | 0.7393 | 0.7945 | 0.944 | 0.9708 | 0.9825 | 0.9834 | 0.9876 | 0.9875 | 0.9958 | 0.9917 |
| ↓ RACC | 0.0542 | 0.063 | 0.0669 | 0.0659 | 0.0651 | 0.0625 | 0.0596 | 0.0628 | 0.063 | 0.0628 | 0.0615 | 0.0628 |
| ↑ TNR | 0.925 | 0.9208 | 0.9069 | 0.9278 | 0.9806 | 0.9903 | 0.9944 | 0.9944 | 0.9958 | 0.9958 | 0.9986 | 0.9972 |
| ↑ TPR | 0.6417 | 0.7708 | 0.7916 | 0.8375 | 0.9833 | 0.9708 | 0.9375 | 0.9875 | 0.9958 | 0.9917 | 0.9792 | 0.9958 |

## 4.4 ROC Curves

Figure 8 presents the ROC curves of the three models on three datasets. It is evident that the EAFP-Med ST, equipped with the EAFP-Med module, demonstrates excellent classification performance across all three datasets.

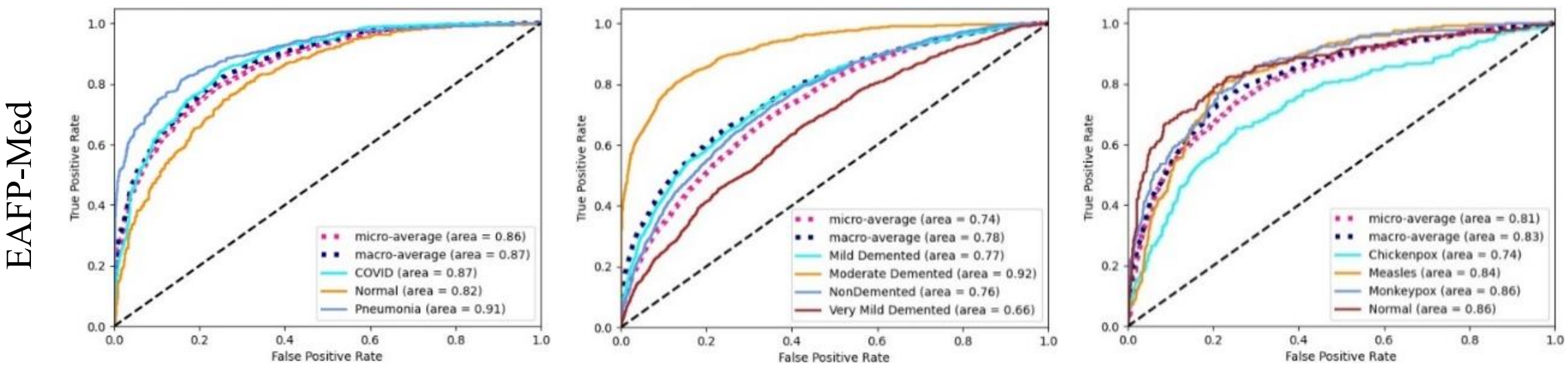

(a) The ROC curve on CXrI. (b) The ROC curve on CMRI. (c) The ROC curve on SK.

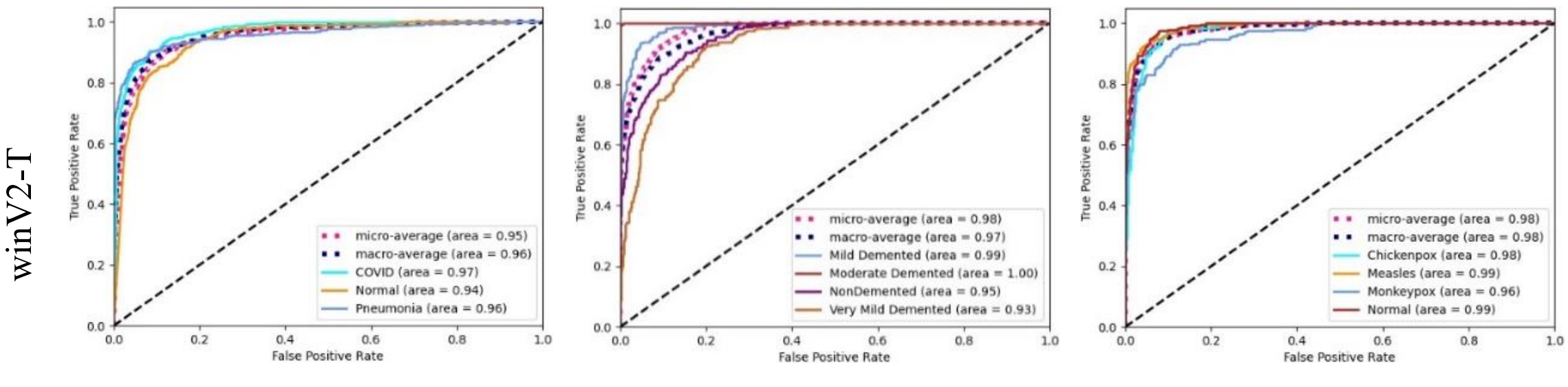

(d) The ROC curve on CXrI. (e) The ROC curve on CMRI. (f) The ROC curve on SK.

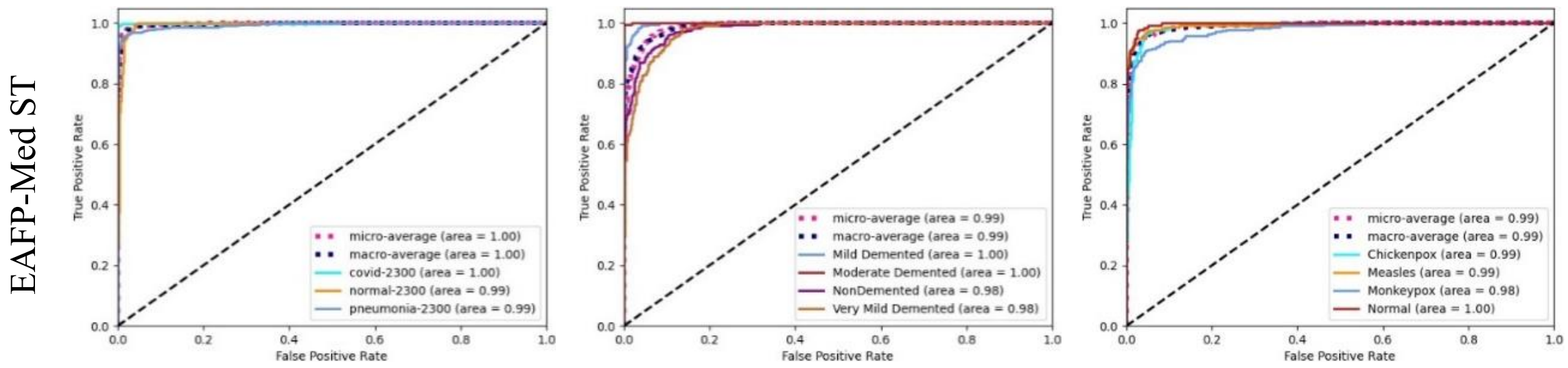

(g) The ROC curve on CXrI. (h) The ROC curve on CMRI. (i) The ROC curve on SK.

**Figure 8 The ROC curves of three methods.**

## 4.5 Grad-CAM ++

Figure 9, Figure 10, and Figure 11, respectively, show the attention visualization results of these methods on three sample images. Grade-CAM++ denotes gradient-weighted class activation mapping plus plus, a visualization technique used to explain model predictions. It generates a class activation map by calculating gradient information to visualize the model's attention to the input image. The GB image refers to the original class activation map, which directly maps gradient information onto the input image to generate a class activation map of the same size as the input image. The CAM-GB image is the result of normalizing the GB image. It smooths the class activation map through a series of operations and better reflects the overall attention of the model to the input image.

Comparing the visualizations of Grade-CAM++, CAM-GB, and GB, it can be proven that EAFP-Med is capable of adaptively and efficiently extracting lesion features from different medical images, providing more effective feature maps to the backbone, enabling the model to accurately locate lesions.

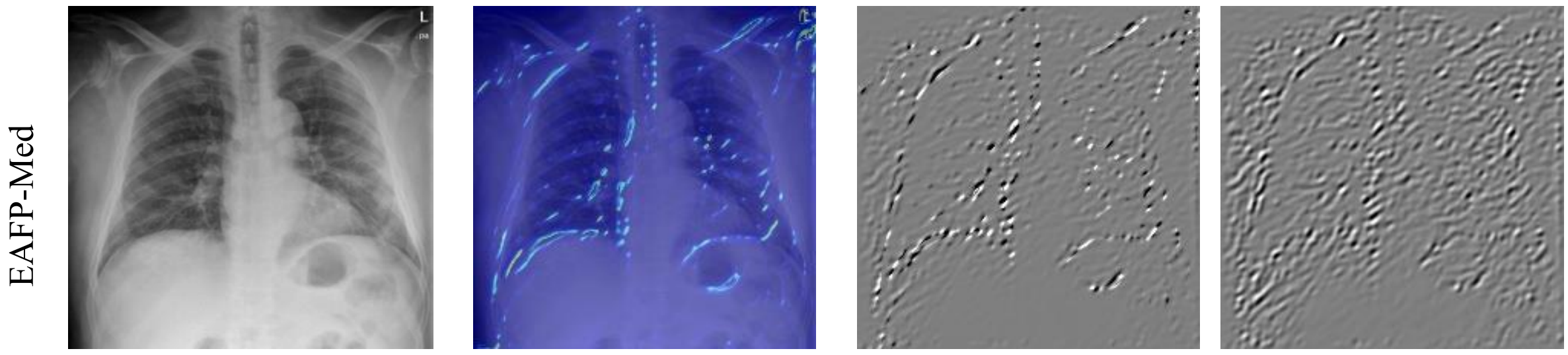

(a) A sample image. (b) Grad-CAM++ image. (c) CAM-GB image. (d) GB image.

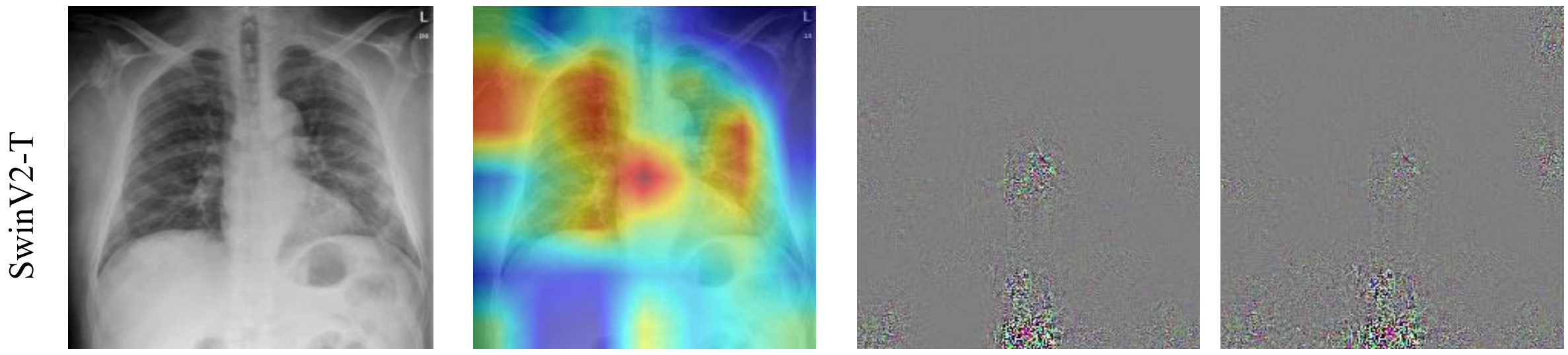

(e) A sample image. (f) Grad-CAM++ image. (g) CAM-GB image. (h) GB image.

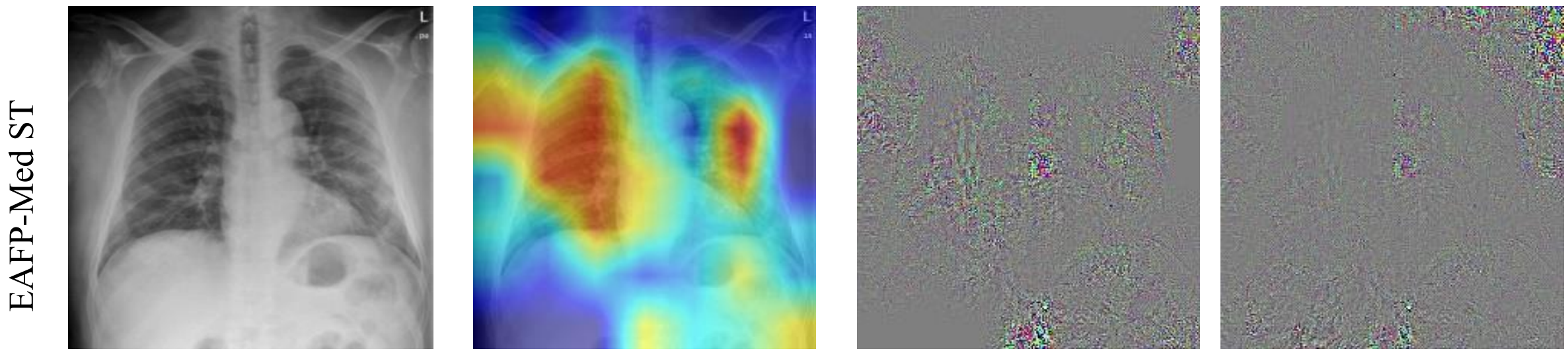

(i) A sample image. (j) Grad-CAM++ image. (k) CAM-GB image. (l) GB image.

**Figure 9 Visualization results of three methods on a sample image from CXrI.**

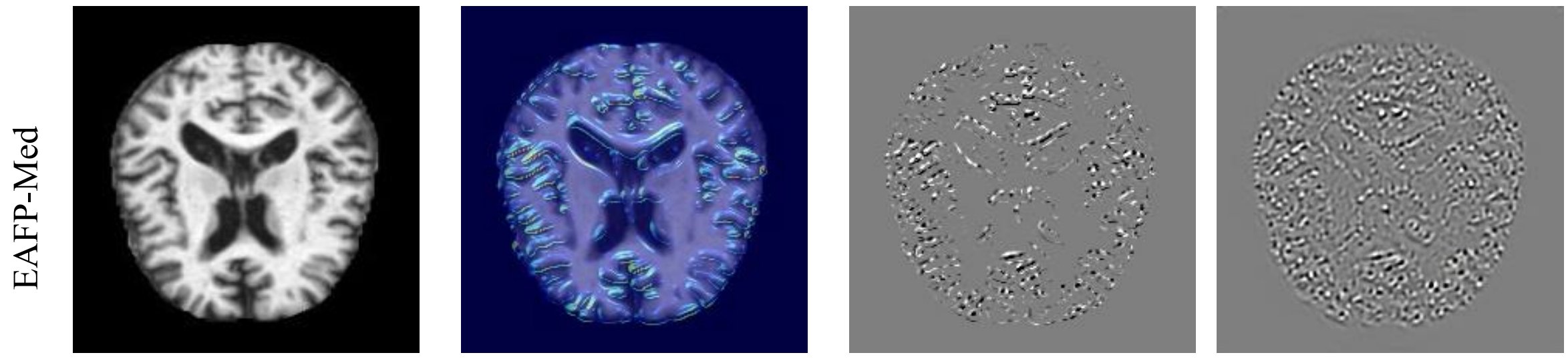

(a) A sample image. (b) Grad-CAM++ image. (c) CAM-GB image. (d) GB image.

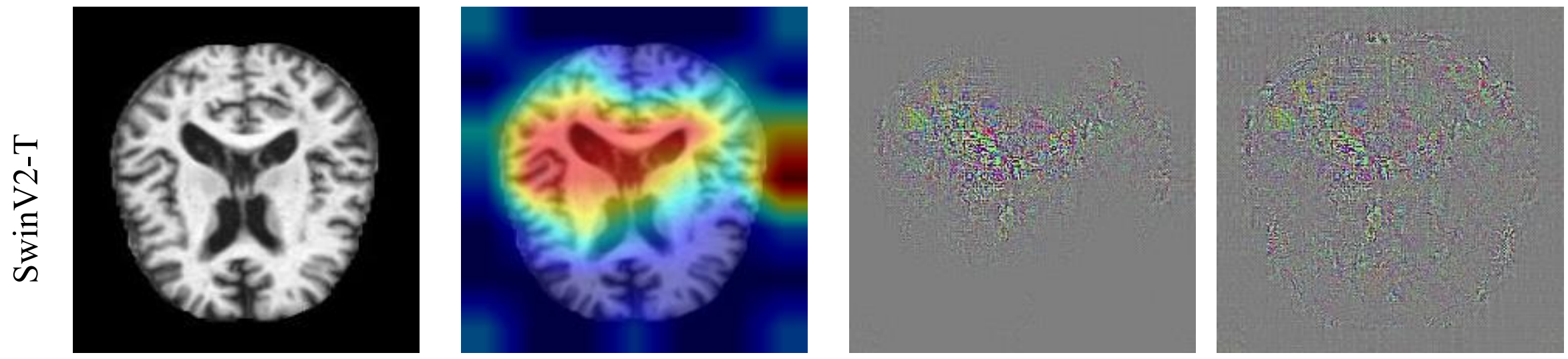

(e) A sample image. (f) Grad-CAM++ image. (g) CAM-GB image. (h) GB image.

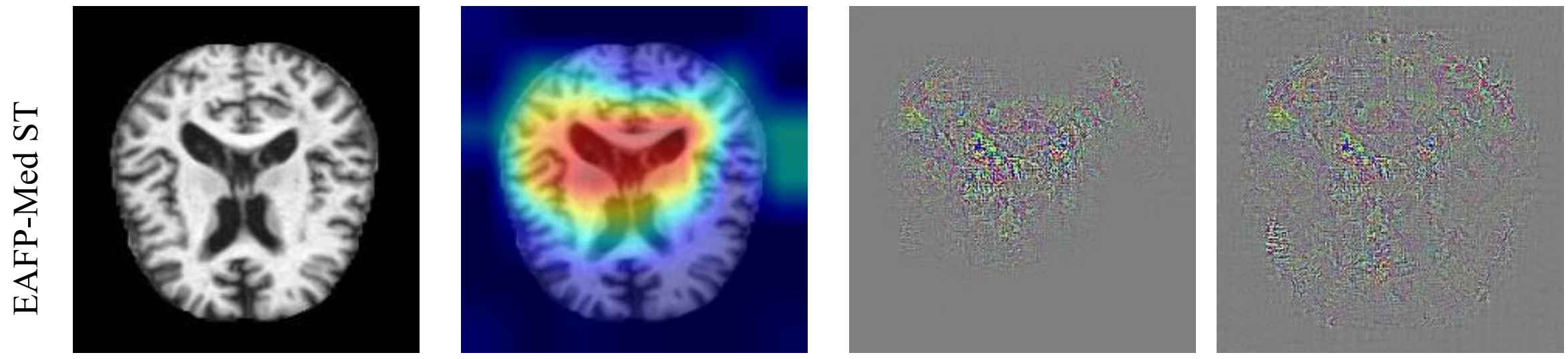

(i) A sample image. (j) Grad-CAM++ image. (k) CAM-GB image. (l) GB image.

**Figure 10 Visualization results of three methods on a sample image from CMRI.**

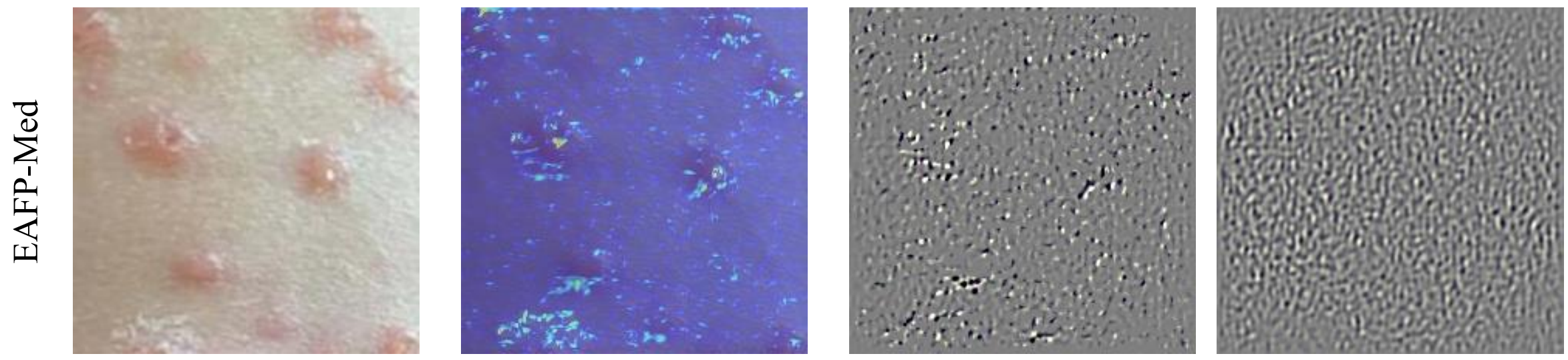

(a) A sample image. (b) Grad-CAM++ image. (c) CAM-GB image. (d) GB image.

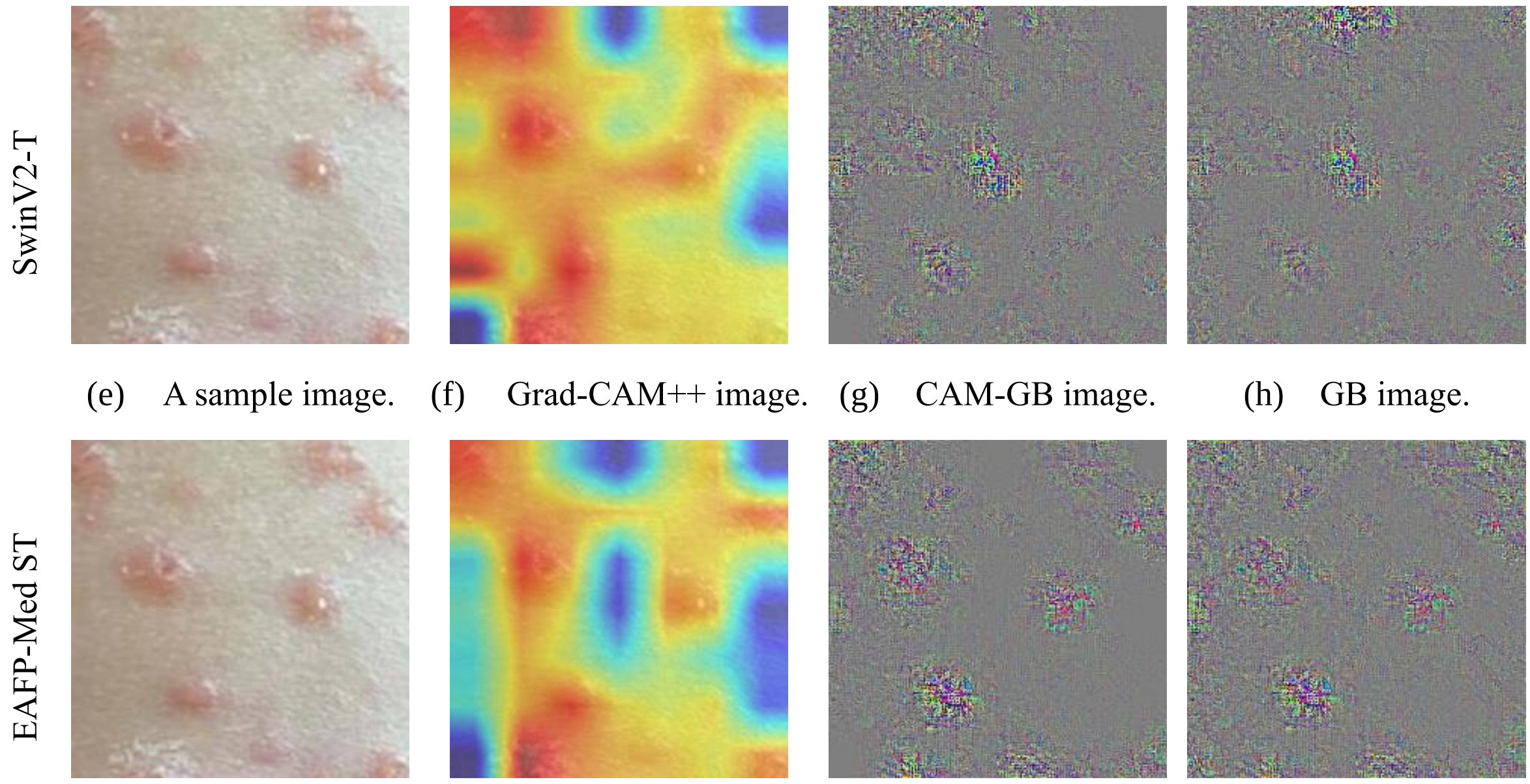

(e) A sample image. (f) Grad-CAM++ image. (g) CAM-GB image. (h) GB image.

(i) A sample image. (j) Grad-CAM++ image. (k) CAM-GB image. (l) GB image.

**Figure 11 Visualization results of three methods on a sample image from SK.**

### 4.6 Comparison with state-of-the-art methods

Table 7 presents the experimental results comparing our method with state-of-the-art methods on three datasets. To measure the performance of each model, we introduced the following five measurement metrics: overall accuracy (OA), false negative rate macro (FNR-M), false positive rate macro (FPR-M), true negative rate macro (TNR-M), overall Matthews correlation coefficient (OM). The experimental results show that EAFP-Med ST achieves state-of-the-art performance in detecting various medical images, demonstrating good adaptability. On the other hand, it also validates the effectiveness of the EAFP-Med module.

**Table 7 Comparison with state-of-the-art methods on three datasets.**

| Datasets | Methods | OA (%) ↑ | FNR-M (%) ↓ | FPR-M (%) ↓ | TNR-M (%) ↑ | OM (%) ↑ |
|---|---|---|---|---|---|---|
| CXrI | DLBD (Alsattar, et al., 2024) | 97.10 | 2.90 | 1.45 | 98.55 | 95.65 |
| | Cal RF+AD+PT (Seethi, et al., 2024) | 97.89 | 2.12 | 1.06 | 98.95 | 96.83 |
| | HiFuse (Huo, et al., 2024) | 97.67 | 2.31 | 1.16 | 98.85 | 96.52 |
| | **EAFP-Med ST (Ours)** | **98.47** | **1.53** | **0.76** | **99.24** | **97.71** |
| CMRI | Conv-eRVFL (Sharma, et al., 2023) | 97.29 | 2.71 | 0.90 | 99.10 | 96.40 |
| | EDCNNS (Lakhan, et al., 2023) | 96.98 | 3.02 | 1.01 | 98.99 | 95.98 |
| | Conv-Swinformer (Hu, et al., 2023) | 97.46 | 2.54 | 0.85 | 99.15 | 96.62 |
| | **EAFP-Med ST (Ours)** | **97.60** | **2.40** | **0.80** | **99.20** | **96.81** |
| SK | MTMS (Akkilic, et al., 2024) | 90.94 | 9.06 | 3.02 | 96.98 | 87.94 |
| | MO-WAE (Asif, et al., 2023) | 98.85 | 1.15 | 0.38 | 99.62 | 98.48 |
| | MPXV-CNN (Thieme, et al., 2023) | 98.22 | 1.77 | 0.59 | 99.41 | 97.65 |
| | **EAFP-Med ST (Ours)** | **99.06** | **0.94** | **0.31** | **99.69** | **98.75** |

## 5 Conclusion

In this paper, we present an efficient adaptive feature processing module based on prompts named EAFP-Med. This module consists of multiple sub-modules (three feature extractors, three feature adaptors, and a prompt-based parameter adaptor), which can adaptively extract lesion features from various medical images based on prompts. By overlaying the extracted features with the initial input images, the lesion features fed to the backbone or downstream network can be enhanced to further improve the diagnosis performance of the model. Moreover, EAFP-Med also has strong applicability, serving as a feature preprocessing module connected to the front end of any backbone, providing a more precise and robust input of lesion features to the downstream network. Experimental results show that the EAFP-Med module can accurately and adaptively extract lesion features from various medical images. Our method, EAFP-Med ST, achieves state-of-the-art performance on all three datasets.

## Appendix

**Table 8 Abbreviation table.**

| Abbreviation | Full Definition |
|---|---|
| ACC | Accuracy. |
| AGF | Adjusted F-score. |
| AGM | Adjusted geometric mean. |
| AUC | Area under the ROC curve. |
| BCD | Bray-Curtis dissimilarity. |
| BM | Informedness or bookmaker informedness. |
| CEN | Confusion entropy. |
| CMRI | Cranial magnetic resonance imaging image dataset. |
| CIA-HGCN | Characteristic information aggregation hypergraph convolutional network. |
| CXrI | Chest X-ray image dataset. |
| DLBD | Dynamic location-based decision. |
| EDCNNS | Evolutionary deep convolutional neural network scheme. |
| F1 | F1 score. |
| FNR | False negative rate. |
| FNR-M | False negative rate macro. |
| FPR | False positive rate. |
| FPR-M | False positive rate macro. |
| G | G-measure geometric mean of precision and sensitivity. |
| Grad-CAM++ | Gradient-weighted class activation mapping plus plus. |
| MCC | Matthews correlation coefficient. |
| MO-WAE | Metaheuristics optimization-based weighted average ensemble model. |
| OA | Overall accuracy. |
| OM | Overall Matthews correlation coefficient. |
| OOC | Otsuka-Ochiai coefficient. |
| PPV | Positive predictive value (Precision). |
| RACC | Random accuracy. |
| ROC | Receiver operating characteristic. |
| SARS-CoV-2 | Severe acute respiratory syndrome coronavirus 2. |
| SK | Skin images dataset. |
| SwinV2-T | Swin Transformer V2 – Tiny. |
| TNR | True negative rate. |
| TPR | True positive rate. |
| WHO | World Health Organization. |

**Acknowledgement**

This work is supported by the China Scholarship Council; the National Natural Science Foundation of China: 91948303-1; the National Key R&D Program of China (2021ZD0140301); the National Natural Science Foundation of China: 611803375, No. 61803375, No. 12002380, No. 62106278, No. 62101575, No. 61906210; the Postgraduate Scientific Research Innovation Project of Hunan Province: QL20210018.

**References**

WHO. "Coronavirus disease (COVID-19)." World Health Organization. https://www.who.int/news-room/fact-sheets/detail/coronavirus-disease-(covid-19) (accessed.

WHO. "Mpox (monkeypox)." World Health Organization. https://www.who.int/health-topics/monkeypox#tab=tab_1 (accessed.

WHO. "2022-23 Mpox (Monkeypox) Outbreak: Global Trends." World Health Organization. https://worldhealthorg.shinyapps.io/mpx_global/ (accessed.

WHO. "Dementia cases set to triple by 2050 but still largely ignored." World Health Organization. https://www.who.int/news/item/11-04-2012-dementia-cases-set-to-triple-by-2050-but-still-largely-ignored (accessed.

WHO. "Dementia: number of people affected to triple in next 30 years." World Health Organization. https://www.who.int/news/item/07-12-2017-dementia-number-of-people-affected-to-triple-in-next-30-years (accessed.

WHO. "WHO reveals leading causes of death and disability worldwide: 2000-2019." World Health Organization. https://www.who.int/news/item/09-12-2020-who-reveals-leading-causes-of-death-and-disability-worldwide-2000-2019 (accessed.

H. A. Alsattar *et al.*, "Developing deep transfer and machine learning models of chest X-ray for diagnosing COVID-19 cases using probabilistic single-valued neutrosophic hesitant fuzzy," (in English), *EXPERT SYSTEMS WITH APPLICATIONS,* vol. 236, FEB 2024, Art no. 121300, doi: 10.1016/j.eswa.2023.121300.

V. D. R. Seethi *et al.*, "An explainable AI approach for diagnosis of COVID-19 using MALDI-ToF mass spectrometry," (in English), *EXPERT SYSTEMS WITH APPLICATIONS,* vol. 236, FEB 2024, Art no. 121226, doi: 10.1016/j.eswa.2023.121226.

X. Z. Huo *et al.*, "HiFuse: Hierarchical multi-scale feature fusion network for medical image classification," (in English), *BIOMEDICAL SIGNAL PROCESSING AND CONTROL,* vol. 87, JAN 2024, Art no. 105534, doi: 10.1016/j.bspc.2023.105534.

X. A. Bi, S. Luo, S. Y. Jiang, Y. Wang, Z. X. Xing, and L. Y. Xu, "Explainable and programmable hypergraph convolutional network for data fusion," (in English), *INFORMATION FUSION,* vol. 100, DEC 2023, Art no. 101950, doi: 10.1016/j.inffus.2023.101950.

R. Sharma, T. Goel, M. Tanveer, P. N. Suganthan, I. Razzak, and R. Murugan, "Conv-eRVFL: Convolutional Neural Network Based Ensemble RVFL Classifier for Alzheimer's Disease Diagnosis," (in English), *IEEE JOURNAL OF BIOMEDICAL AND HEALTH INFORMATICS,* vol. 27, no. 10, pp. 4995-5003, OCT 2023, doi: 10.1109/JBHI.2022.3215533.

A. Lakhan, T. M. Gronli, G. Muhammad, and P. Tiwari, "EDCNNS: Federated learning enabled evolutionary deep convolutional neural network for Alzheimer disease detection," (in English), *APPLIED SOFT COMPUTING,* vol. 147, NOV 2023, Art no. 110804, doi: 10.1016/j.asoc.2023.110804.

Z. Hu, Y. Li, Z. Wang, S. Zhang, and W. Hou, "Conv-Swinformer: Integration of CNN and shift window attention for Alzheimer's disease classification," *Computers in Biology and Medicine,* vol. 164, p. 107304, 2023/09/01/ 2023, doi: https://doi.org/10.1016/j.compbiomed.2023.107304.

A. N. Akkilic, Z. Sabir, S. A. Bhat, and H. Bulut, "A radial basis deep neural network process using the Bayesian regularization optimization for the monkeypox transmission model," (in English), *EXPERT SYSTEMS WITH*

*APPLICATIONS,* vol. 235, JAN 2024, Art no. 121257, doi: 10.1016/j.eswa.2023.121257.

S. Asif, M. Zhao, F. X. Tang, Y. S. Zhu, and B. K. Zhao, "Metaheuristics optimization-based ensemble of deep neural networks for Mpox disease detection," (in English), *NEURAL NETWORKS,* vol. 167, pp. 342-359, OCT 2023, doi: 10.1016/j.neunet.2023.08.035.

A. H. Thieme *et al.*, "A deep-learning algorithm to classify skin lesions from mpox virus infection," (in English), *NATURE MEDICINE,* vol. 29, no. 3, pp. 738-+, MAR 2023, doi: 10.1038/s41591-023-02225-7.

Y. Li *et al.*, *Efficient and Explicit Modelling of Image Hierarchies for Image Restoration*. 2023.

A. Chattopadhay, A. Sarkar, P. Howlader, and V. N. Balasubramanian, "Grad-CAM++: Generalized Gradient-Based Visual Explanations for Deep Convolutional Networks," in *2018 IEEE Winter Conference on Applications of Computer Vision (WACV)*, 12-15 March 2018 2018, pp. 839-847, doi: 10.1109/WACV.2018.00097.

A. Asraf. *COVID19 Pneumonia Normal Chest Xray PA Dataset*. [Online]. Available: https://www.kaggle.com/datasets/amanullahasraf/covid19-pneumonia-normal-chest-xray-pa-dataset

Uraninjo. *Augmented Alzheimer MRI Dataset V2*. [Online]. Available: https://www.kaggle.com/datasets/uraninjo/augmented-alzheimer-mri-dataset-v2

D. Bala. *Monkeypox Skin Images Dataset (MSID)*. [Online]. Available: https://www.kaggle.com/datasets/dipuiucse/monkeypoxskinimagedataset